\documentclass[11pt,a4paper]{article}

\usepackage{graphicx}
\usepackage{amsmath}
\usepackage{amsfonts}
\usepackage{amssymb}
\usepackage{enumerate}
\usepackage{lscape}
\usepackage{longtable}
\usepackage{rotating}
\usepackage{color}
\usepackage{url}
\usepackage[font={small,it}]{caption}
\usepackage{subcaption}
\usepackage{rotating}
\usepackage{titlesec}
\usepackage[title]{appendix}

\usepackage{mathrsfs}
\usepackage{upgreek}

\usepackage{algorithmicx}
\usepackage{algorithm}
\usepackage[noend]{algpseudocode}

\newtheorem{theorem}{Theorem}[section]

\newtheorem{definition}{Definition}[section]

\makeatletter
\newcommand*{\centerfloat}{%
  \parindent \z@
  \leftskip \z@ \@plus 1fil \@minus \textwidth
  \rightskip\leftskip
  \parfillskip \z@skip}
\makeatother

\newcommand{\DI}{\text{DI}}
\newcommand{\CV}{\text{CV}}

\usepackage{soul}

\numberwithin{equation}{section}
\definecolor{forestgreen}{rgb}{0.13, 0.55, 0.13}
\definecolor{greenp}{rgb}{0.0, 0.65, 0.31}
\definecolor{greenr}{rgb}{0.4, 0.69, 0.2}
\definecolor{indiagreen}{rgb}{0.07, 0.53, 0.03}
\definecolor{ffqqqq}{rgb}{0.082,0.40,0.75}

\usepackage[square,numbers]{natbib}
\usepackage{hyperref}
\hypersetup{
  colorlinks,
  citecolor=indiagreen,
  linkcolor=indiagreen,
  urlcolor=ffqqqq}

\begin{document}

\title{The Sharpe predictor for fairness in machine learning}

\author{
S. Liu\thanks{Department of Industrial and Systems Engineering,
Lehigh University,
200 West Packer Avenue, Bethlehem, PA 18015-1582, USA
({\tt sul217@lehigh.edu}).}
\and
L. N. Vicente\thanks{Department of Industrial and Systems Engineering,
Lehigh University,
200 West Packer Avenue, Bethlehem, PA 18015-1582, USA. Support for this
author was partially provided by the Centre for Mathematics
of the University of Coimbra under grant FCT/MCTES
UIDB/MAT/00324/2020.}
}

\maketitle
\footnotesep=0.4cm
{\small
\begin{abstract}
In machine learning (ML) applications, unfair predictions may discriminate against a minority group. Most existing approaches for fair machine learning (FML) treat fairness as a constraint or a penalization term in the optimization of a ML model, which does not lead to the discovery of the complete landscape of the trade-offs among learning accuracy and fairness metrics, and does not integrate fairness in a meaningful way.

Recently, we have introduced a new paradigm for FML based on Stochastic Multi-Objective Optimization (SMOO), where accuracy and fairness metrics stand as conflicting objectives to be optimized simultaneously. The entire trade-offs range is defined as the Pareto front of the SMOO problem, which can then be efficiently computed using stochastic-gradient type algorithms. SMOO also allows defining and computing new meaningful predictors for FML, a novel one being the Sharpe predictor that we introduce and explore in this paper, and which gives the highest ratio of accuracy-to-unfairness. Inspired from SMOO in finance, the Sharpe predictor for FML provides the highest prediction return (accuracy) per unit of prediction risk (unfairness).
\end{abstract}
}

\section{Introduction}


Machine learning algorithms have an increasingly impact on real-life decision-making systems in our society. In many critical applications including credit-scoring, healthcare, hiring, and criminal justice, it is paramount to guarantee that the prediction outcome is both accurate and fair with respect to sensitive attributes such as gender and race. A collection of literature work on algorithm fairness in classification, regression, and clustering problems (see e.g. \cite{RBerk_etal_2017, RBerk_etal_2021, TCalders_etal_2013, FChierichetti_etal_2017, SCorbett-Davies_etal_2017, MHardt_EPrice_NSrebro_2016, JKleinberg_SMullainathan_MRaghavan_2016, JKomiyama_etal_2018, BWoodworth_etal_2017, MBZafar_etal_2017, MBZafar_etal_2017b, RZemel_etal_2013}) focuses on either defining fairness or developing fair learning algorithms. Being of conflicting nature, accuracy and fairness give rise to a set of trade-offs predictors, which can be calculated using stochastic bi-objective optimization methods (penalization~\cite{NMartinez_MBertran_GSapiro_2019, NMartinez_MBertran_GSapiro_2020}, epsilon-constrained~\cite{MBZafar_etal_2017, MBZafar_etal_2017b}, stochastic bi-gradient~\cite{SLiu_LNVicente_2019, SLiu_LNVicente_2020, WRMonteiro_GReynoso-Meza_2021}). Each of such predictors is non-dominated in the sense that there is no other one simultaneously more accurate and fair. The trade-offs can be typically described by a curve on the accuracy-fair plane, the so-called Pareto front.


In this paper we deal with bi-objective optimization problems of the form $\min F(x)$, where $F: \mathbb{R}^n \rightarrow \mathbb{R}^2$ is a vector function $F=(f_1,f_2)^\top$ defined by two component functions~$f_1$ and~$f_2$. When~$f_1$ and~$f_2$ are conflicting, no point exists that minimizes both simultaneously.
The notion of Pareto dominance allows for a comparison between any two points and leads to the definition of Pareto optimality.
One says that~$x$ dominates~$y$ if $F(x) < F(y)$ componentwise. A point~$x$ is then called a Pareto minimizer if it is not dominated by any other one.
The so-called Pareto front (or efficient frontier) $F(\mathcal{P})$ is formed by mapping all elements of $\mathcal{P}$ into the decision space~$\mathbb{R}^2$,
$F(\mathcal{P})= \{F(x): x \in \mathcal{P}\}$.

Several recent works~\cite{SLiu_LNVicente_2019, SLiu_LNVicente_2020, SLiu_LNVicente_2021, NMartinez_MBertran_GSapiro_2020, MBZafar_etal_2017, MBZafar_etal_2017b} attempt to capture a complete trade-off between fairness and prediction or clustering quality. In particular, \cite{SLiu_LNVicente_2019, SLiu_LNVicente_2020, SLiu_LNVicente_2021} aimed at constructing an entire Pareto front using a posteriori approaches, whereas \cite{NMartinez_MBertran_GSapiro_2020, MBZafar_etal_2017, MBZafar_etal_2017b} illustrated the trade-offs with a portion of Pareto front computed by a priori methods where decision-makers' preferences are specified before optimization. The whole Pareto front is not easy to compute especially when the function and gradient evaluations are computationally expensive. Moreover, after obtaining the Pareto front, there is still lack of guideline for the decision-makers to select appropriate solutions. To the best of our knowledge, the existing principles for selecting a certain accuracy-fairness trade-off include the following: 
\begin{itemize}
    \item Given the weights associated with each objective, solving a corresponding weighted-sum problem provides a potential solution candidate.
    \item  Maximizing prediction accuracy~\cite{MBZafar_etal_2017, MBZafar_etal_2017b} subject to the amount of fairness greater than a desired threshold. 
    \item Maximizing the worst performance across sensitive groups~\cite{EDiana_etal_2021, NMartinez_MBertran_GSapiro_2020} leads to lowest performance disparity. 
\end{itemize}

We introduce a new accuracy/fairness trade-off concept inspired from the definition of the Sharpe ratio~\cite{WFSharpe_1966} developed in the sixties for portfolio selection in finance (see, e.g.,~\cite[Section~8.2]{GCornneujols_RTuntuncu_2007}). A new predictor, called Sharpe, is defined as the one that provides the highest gain in accuracy return per unit of risk fairness, thus providing a single trade-off with a meaningful property on the Pareto front. The Sharpe predictor depends on the fairness definition used, and each one will lead to a different rate of accuracy or misclassification.
The use of the Sharpe predictor will thus allow the comparison of different fairness concepts in fair machine learning.
To our knowledge this is the first attempt to identify a prediction data point that could be put in relation to the accuracy/fairness Pareto front, rather than the two front extreme points corresponding to highest accuracy and fairness or any process involving the artificial selection of a weight or a threshold parameter.

\section{Fairness in supervised machine learning}
\label{fairness_introduction}

In supervised machine learning, training samples consist of pairs of feature vectors (containing a number of features that are descriptive of each instance) and target values/labels. One tries to use the training data to find a predictor, seen as a function mapping feature vectors into target labels, that is accurate on testing data. Such a predictor is typically characterized by a number of parameters, and the process of identifying the optimal parameters is called training or learning. The trained predictor can then be used to predict the labels of test instances.
If a ML predictor inequitably treats samples from different groups defined by \textit{sensitive} or \textit{protected} attributes, such as gender and race, one says that such a predictor is \textit{unfair}.
Simply excluding sensitive attributes from the features (called \textit{fairness through unawareness}) does not help as the sensitive attributes (e.g., race) can be inferred from the remaining features (such as the highest educational degree)~\cite{DPedreshi_SRuggieri_FTurini_2008, RZemel_etal_2013}.

Depending on when the fairness criteria are imposed, there are three categories of approaches proposed to handle fairness, namely pre-processing, in-training, and post-processing.
Pre-processing approaches~\cite{FCalmon_etal_2017, RZemel_etal_2013} transform the input data representation so that the predictions from any model trained on the transformed data are guaranteed to be fair. Post-processing~\cite{MHardt_EPrice_NSrebro_2016, GPleiss_etal_2017} tries to adjust the output of a trained predictor to increase fairness while maintaining the prediction accuracy as much as possible.
Even with fairness through unawareness, fairness can be enforced by in-training methods: If the sensitive attributes are accessible in the training samples,
in-training methods~\cite{SBarocas_ADSelbst_2016, TCalders_FKamiran_MPechenizkiy_2009, TKamishima_SAkaho_JSakuma_2011, MBZafar_etal_2017b, BWoodworth_etal_2017, MBZafar_etal_2017} enforce fairness during the training process either by directly solving fairness-constrained optimization problems, or by adding fairness penalty terms to the learning objective.
The approach introduced in this paper falls into the in-training category.

Some of the broadest fairness criteria are~\textit{disparate impact}~\cite{SBarocas_ADSelbst_2016} (also called~\textit{demographic parity}~\cite{TCalders_FKamiran_MPechenizkiy_2009}), \textit{Overall accuracy equality}~\cite{RBerk_etal_2021}, \textit{equalized odds}, and its special case of \textit{equal opportunity}~\cite{MHardt_EPrice_NSrebro_2016}, corresponding to different aspects of fairness. Since many real-world decision-making problems, such as college admission, bank loan application, hiring decisions, etc., can be formulated as binary classification tasks, we
will focus in this paper on the binary classification setting. Let $Z \in \mathbb{R}^n$, $A \in \{0, 1\}$, and $Y \in \{-1, + 1\}$ respectively denote feature vector, binary-valued sensitive attribute,
and target label. Consider a general predictor $\hat{Y} \in \{-1, + 1\}$ which depends on $Z$, $A$, or both.  The predictor is free of disparate impact~\cite{SBarocas_ADSelbst_2016} if the prediction outcome is independent of the sensitive attribute, i.e., for $\hat{y} \in \{-1, +1\}$,
\begin{equation}
\label{def_disparate_impact}
\mathbb{P}\{\hat{Y} = \hat{y} | A = 0\} \;=\; \mathbb{P}\{\hat{Y} = \hat{y} | A = 1\}.
\end{equation}
Disparate impact may be be unrealistic when one group is more likely to be classified as a positive class than the other~\cite{JKelly_forbes_2020}. In such a case, one can use equalized odds~\cite{MHardt_EPrice_NSrebro_2016} (requiring conditionally independence of the sensitive attribute given the true outcome) and equal opportunity~\cite{MHardt_EPrice_NSrebro_2016} (a relaxation of equalized odds imposing such independence only for instances with positive target value). More fairness and bias measures are well-explained in~\cite{SBarocas_MHardt_ANarayanan_2017, Mehrabi2019ASO, SVerma_JRubin_2018}.

\section{The Sharpe predictor for fairness in machine learning}
\label{sec:PW1}

\begin{figure}{r}
  \centering
  \includegraphics[width = 4.5cm]{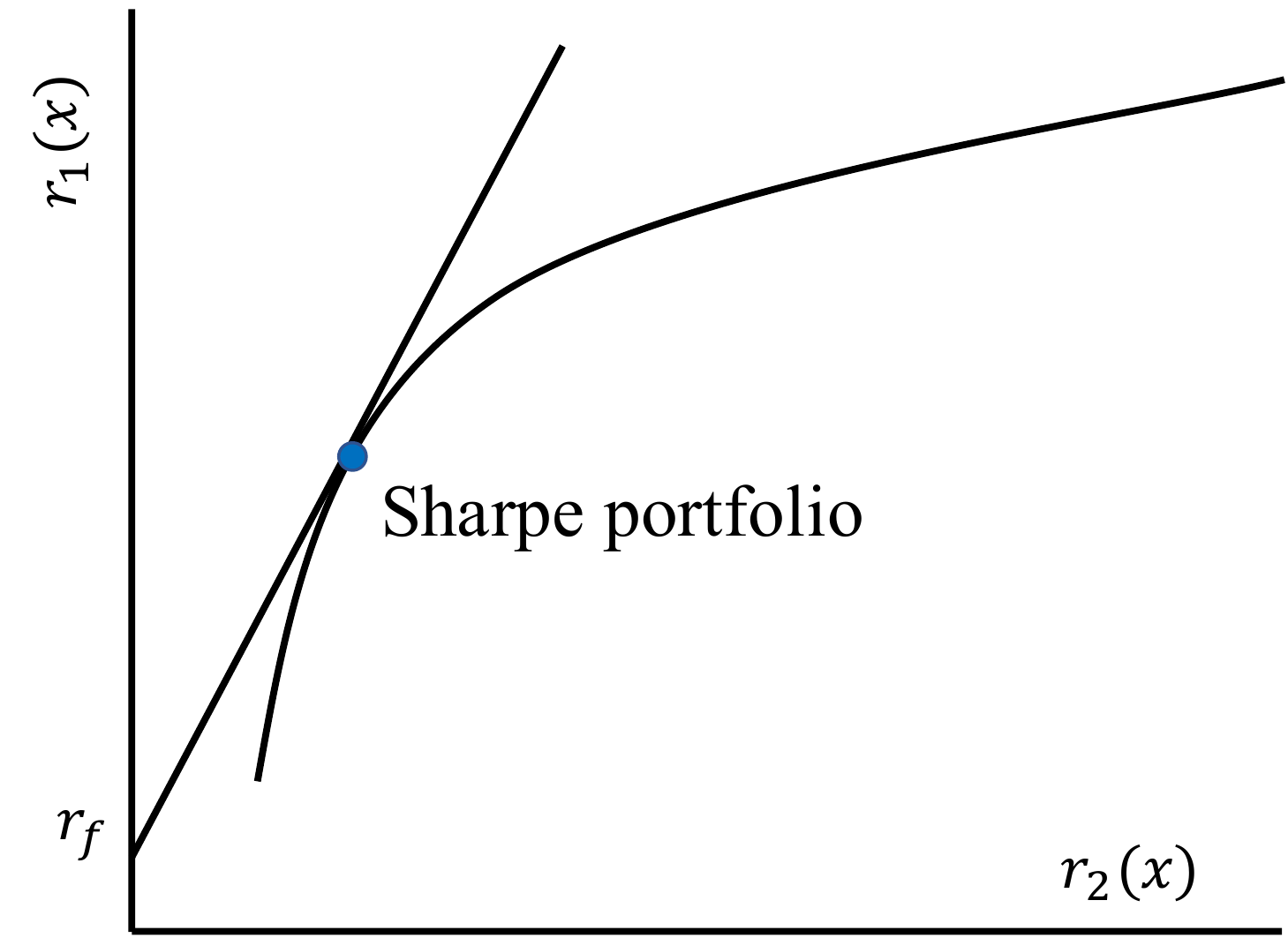}
  \caption{Illustration of Sharpe ratio in portfolio selection.
  \label{fig:sharpe_ratio}}
\end{figure}

The Markowitz selection of finance portfolios is done by trading-off return and risk. A portfolio is defined by a vector~$x \in X$ denoting the proportions invested in each portfolio asset. Let~$R$ denote a vector containing the stochastic returns of all assets. The expected return of the portfolio~$x$ is given by $r_1(x)=\mathbb{E}(R)^\top x$.
The risk of the portfolio~$x$ is given by the variance of its return, $r_2(x)=x^\top Q x$, where~$Q$ is the covariance matrix of~$R$. Functions $r_1$ and $r_2$ are conflicting. A popular way to determine a portfolio on the Pareto front with desirable properties is by maximizing the so-called Sharpe ratio $(r_1(x) - r_f)/r_2(x)$, where $r_f$ is the expected return of a riskless portfolio ($r_f < \min_{x \in X} r_1(x)$). The Sharpe portfolio gives the highest
reward-to-variability ratio or the largest differential of expected return by unit of risk (relatively to a risk-free portfolio). See Figure~\ref{fig:sharpe_ratio}.
Sharpe portfolios have other desirable features, in particular being well diversified in their asset composition.

We are proposing, in the context of fairness in ML, the definition and computation of a {\it Sharpe predictor}. Prediction accuracy  will be called {\it prediction return}. The lowest the loss or training error $f_1(x)$ is, the highest the return will be.
Prediction unfairness will be called {\it prediction risk}. The highest the fairness measure $f_2(x)$ is, the highest the risk will be.
The Sharpe predictor is the one which gives us the highest prediction return per unit of prediction risk (relatively to a risk-free predictor), see Definition~\ref{def:Sharpe}.

\begin{definition} \label{def:Sharpe}
The Sharpe predictor $x_S$ is calculated by solving the single-level stochastic optimization problem
\begin{equation} \label{Sharpe-predictor}
\max_x \;
\frac{f_f-f_1(x)}{f_2(x)} \;=\; \frac{-f_1(x)-(-f_f)}{f_2(x)},
\end{equation}
where $-f_1$ is the prediction return ($f_1$ is the prediction loss), $f_2$ is the prediction risk, and $f_f > \max_x f_1(x)$ is the prediction loss of a risk-free predictor.
\end{definition}

The Sharpe predictor is a Pareto minimizer or non-dominated solution of the bi-objective optimization problem $\min (f_1,f_2)$.

\begin{theorem}
The Sharpe predictor $x_S$ is a Pareto minimizer of $\min (f_1,f_2)$.
\end{theorem}

The proof is elementary and known in the literature Sharpe portfolio selection. If $x_S$ is dominated, there exists a $x_d$ such that $f_1(x_d) < f_1(x_S)$ and $f_2(x_d) < f_2(x_S)$. A contradiction would arise because
$(f_f-f_1(x_d)) / f_2(x_d) > (f_f-f_1(x_S)) / f_2(x_S)$.

Given any predictor~$x$, one has $(f_f-f_1(x_S)) / f_2(x_S) > (f_f-f_1(x)) / f_2(x)$
(equality can only occur if one of the functions is not strictly convex).
Hence, $f_2(x) / f_2(x_S) > (f_f-f_1(x)) / (f_f-f_1(x_S))$, which means that if the prediction risk increases from $x_S$ to~$x$, such a change is always larger than the one in prediction return (relative to the return of a risk-free prediction).

\section{The Sharpe predictor for accuracy vs disparate impact}
\label{results_DI}

Disparate impact (\ref{def_disparate_impact}) is one of the most commonly used fairness criteria in supervised machine learning.
A general measurement of disparate impact, the so-called CV score~\cite{TCalders_SVerwer_2010}, is defined by the maximum gap between the probabilities of getting positive outcomes in different sensitive groups, i.e.,
\begin{equation} \label{CVscore}
    \CV(\hat{Y}) \;=\; |\mathbb{P}\{\hat{Y} = 1 | A = 0\} - \mathbb{P}\{\hat{Y} = 1| A = 1\}|.
\end{equation}
where the binary predictor $\hat{Y} = \hat{Y}(Z; x) \in \{-1, +1\}$ depends on $x$ and maps from $Z$ to $Y$.
As suggested in~\cite{,SLiu_LNVicente_2020},
the trade-offs between prediction accuracy and disparate impact can then be formulated as the following stochastic bi-objective optimization problem:
\begin{equation}
      \min \; ( f_1(x),f_2(x) ) = \left( \mathbb{E}[\ell(\hat{Y}(Z; x), Y), \CV(\hat{Y}(Z; x)) \right).
    \label{obj1_loss}
\end{equation}
The first objective in~\eqref{obj1_loss} is a composition function of a loss function $\ell(\cdot,\cdot)$ and the prediction function $\hat{Y}(Z; x)$, and the expectation is taken over the joint distribution of $Z$ and $Y$.

Let us consider logistic regression model for binary classification, where the parameters are $x=(c,b)^\top$, with $c\in\mathbb{R}^n$ and $b$ a scalar. We assume that $N$ \textit{independent} samples $\{z_j, a_j, y_j\}_{j = 1}^N$ are given.
For a given feature vector $z_i$ and the corresponding true label $y_i$, one searches for a separating hyperplane $\phi(z_j; x) = \phi(z_j; c, b) = c^\top z_j + b$ such that
$c^\top z_j + b  \geq (<) 0$ when $y_j = +1(-1)$, leading to the threshold classifier $\hat{Y}(z_j; c, b) = 2 \times \mathbb{1}(c^\top z_j + b \geq 0) - 1$.
The logistic regression loss $\ell(z, y; c, b) = \text{log}(1+\text{exp}(-y(c^\top z + b)))$ is a smooth and convex surrogate of the $0$-$1$ loss.

The first objective in~\eqref{obj1_loss} can then be approximated by the empirical logistic regression loss, i.e.,
\[
    f_1(c, b) \;=\; \tfrac{1}{N}\sum_{j=1}^N \text{log}(1+\text{exp}(-y_j(c^\top z_j + b))),
\]
based on $N$ independent training samples. A regularization term $(\lambda/2) \|c\|^2$ can be added to avoid over-fitting.
The second objective is to minimize $\CV(\hat{Y}(Z; x))$. In~\cite{MBZafar_etal_2017b}, it was proposed to avoid the non-convexity of the second objective using \textit{decision boundary covariance} as a convex approximate measurement of disparate impact.
Specifically, the CV score~(\ref{CVscore}) can be approximated by the empirical covariance between the sensitive attributes $A$ and the hyperplane $\phi(Z; c, b)$,
\begin{equation*}
 \mbox{Cov}(A, \phi(Z; c, b))
       \;= \;  \mathbb{E}[(A - \bar{A})\phi(Z; c, b)] - \mathbb{E}[A - \bar{A}]\bar{\phi} (Z; c, b)
      \;\simeq\; \tfrac{1}{N}\textstyle\sum_{j = 1}^N (a_j - \bar{a})\phi(z_j; c, b),
\label{Approx_def_disparate_impact}
\end{equation*}
where $\bar{A}$ is the expected value of the sensitive attribute, and $\bar{a}$ is an approximated value of~$\bar{A}$ using $N$ samples.
The intuition behind this approximation is that the disparate impact~\eqref{def_disparate_impact} basically requires the predictor to be completely independent of the sensitive attribute. Given that zero covariance is a necessary condition for independence, the second objective can be approximated as:
\[
\textstyle
f_2^\DI(c, b) \;=\; [\frac{1}{N}\sum_{j=1}^N (a_j - \bar{a})(c^\top z_j + b)]^2,
\]
which is monotonically increasing with disparate impact. We construct the finite-sum bi-objective problem
\begin{equation}
    \min \; \left(f_1(c, b), f_2^\DI(c, b)\right),
    \label{bi_obj_DI_binary}
\end{equation}
where both objective functions are convex and smooth.

\begin{figure}{r}
  \centering
 \includegraphics[width = 5.5cm]{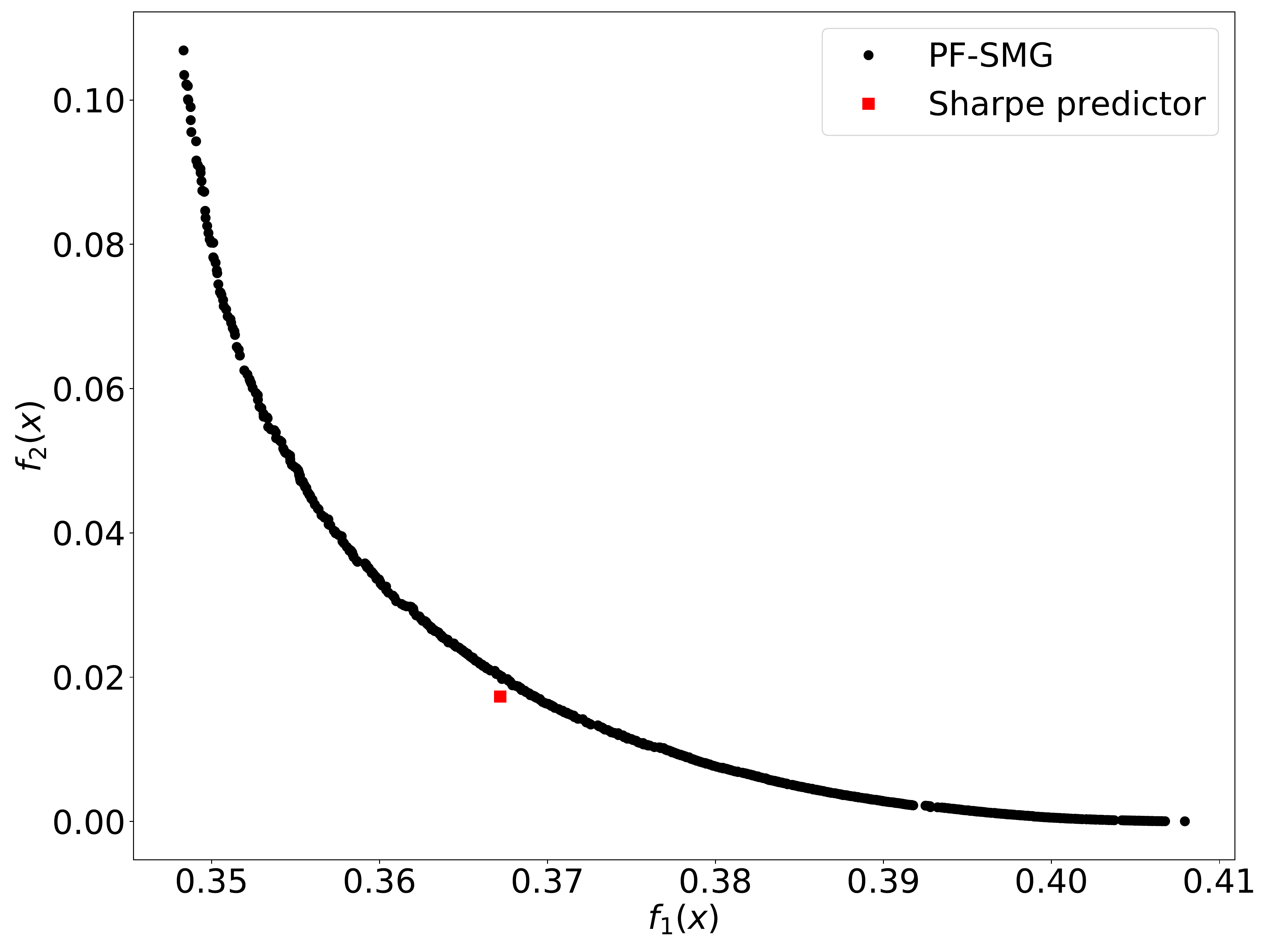}
 \includegraphics[width = 5.5cm]{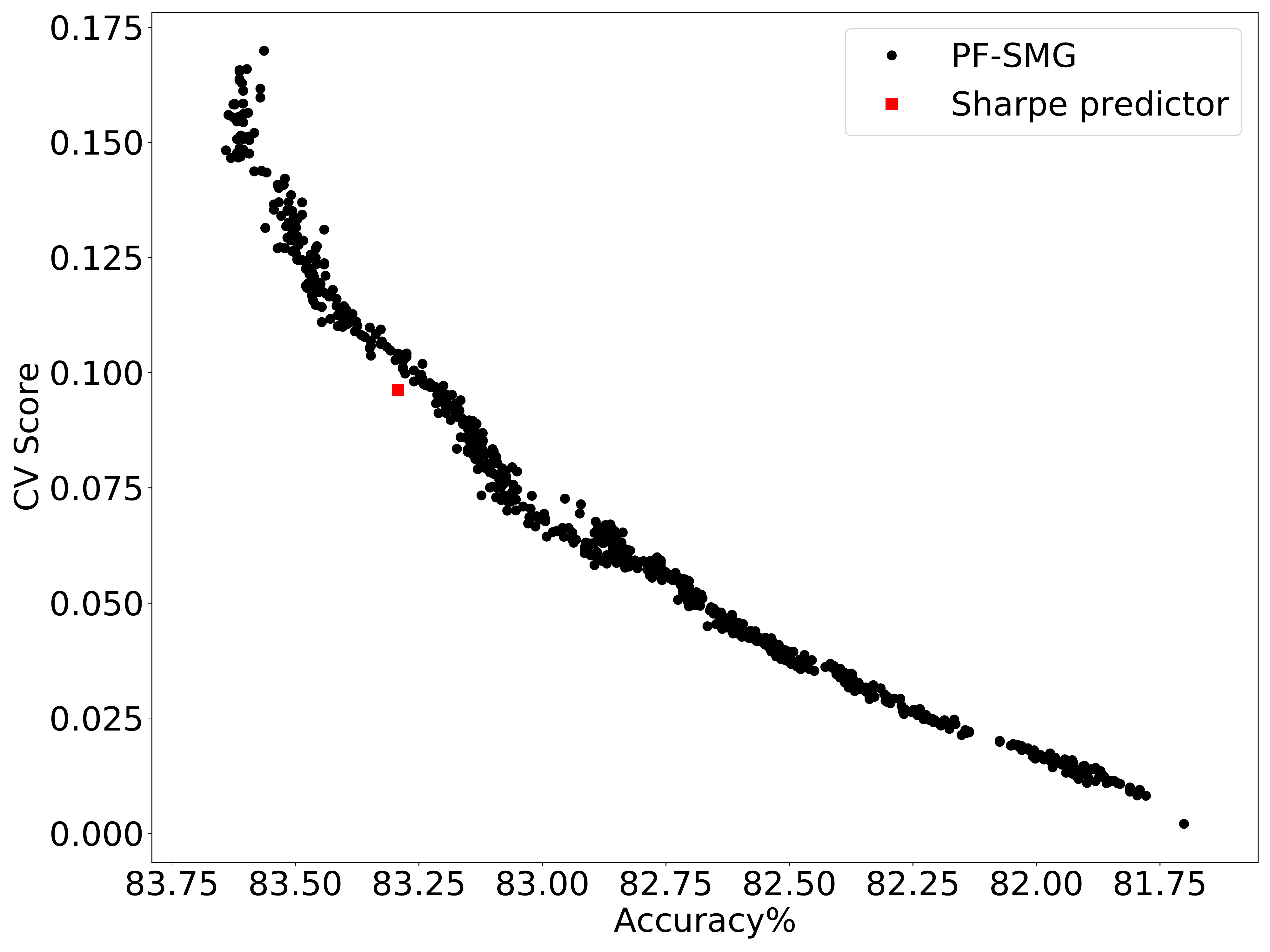}
  \caption{Sharpe predictor and the Pareto front using Adult Income dataset w.r.t. gender.
  The parameter $f_f$ in~(\ref{Sharpe-predictor}) was set to $0.37$.
  Parameters used in PF-SMG algorithm~\cite{SLiu_LNVicente_2019}: step size $0.01$ and number of iterations $6,500$.}
  \label{fig:sharpe_predictor}
\end{figure}

For the illustration of the Sharpe predictor, we will use the experiments reported in~\cite{SLiu_LNVicente_2020}, where we used the \textit{Adult Income} dataset~\cite{RKohavi_1996} from the UCI Machine Learning Repository~\cite{DDua_and_CGraff_2017}. The cleaned up version of \textit{Adult Income} dataset contains 45,222 samples. Each instance is characterized by 12 nonsensitive attributes (including age, education, marital status, etc.) and a binary sensitive attribute (gender). The goal is to predict whether a person makes over 50K per year. The sensitive attribute chosen was gender. In~\cite{SLiu_LNVicente_2020}, we have randomly chosen 5,000 training instances, using the remaining instances as the testing set for presenting the results. The PF-SMG algorithm~\cite{SLiu_LNVicente_2019} was applied to solve~(\ref{bi_obj_DI_binary}), meaning to compute its Pareto front.
The computed Pareto fronts are given in Figure~\ref{fig:sharpe_predictor}, from which we can see the conflicting nature of the two objectives.
Figure~\ref{fig:sharpe_predictor} (left) shows $f_1$ (prediction error) vs $f_2^\DI$ (empirical covariance).
Figure~\ref{fig:sharpe_predictor} (right) shows accuracy vs disparate impact (CV score).

The ranges of $(f_1,f_2)$ in ML are not of the same nature of those of $(r_1,r_2)$ in finance, and one difference forces us to consider $f_f$ as the accuracy of a low-risk predictor (instead of a risk-free one). The value of $f_f$ should be selected lower than $\max_x f_1(x)$ but relatively close to it. For example, in order to compute a Sharpe predictor for the bi-objective problem~\eqref{bi_obj_DI_binary}, we pose the minimization problem $\min~(f_1(c, b) - f_f)/f_2^{\DI}(c, b)$, and solve it using the stochastic gradient method. 

One can compute a biased stochastic gradient using a batch of samples, as $f_1$ and $f_2^{\DI}$ have finite-sum forms. A gradient estimator $g^N$ is said to be consistent~\cite{JChen_RLuss_2018} if for any $\epsilon > 0$, $\lim_{N \rightarrow \infty} \mathbb{P}(\|g^N - \nabla f\| > \epsilon) = 0$ holds, where $\nabla f$ is the true gradient. Such a consistent gradient estimator can be computed in our implementation by increasing the batch size along the iterations, obtaining increasingly accurate gradients. In our case, where both objectives are convex, an $\mathcal{O}(1/\sqrt{T})$ convergence rate of stochastic gradient descent was proved for consistent but biased gradient estimators (see~\cite{JChen_RLuss_2018} for more details). 

Using the same experiment setting as for the Pareto front calculation, we set $f_f = 0.37$,  which is slightly less than the maximum training error ($0.394$) when using $5,000$ training samples. In Figure~\ref{fig:sharpe_predictor}, the testing error, CV score, and testing accuracy using testing samples are evaluated for the obtained Sharpe predictor and marked by a red square. As expected, the Sharpe predictor is on the Pareto front.

\section{The Sharpe predictor more broadly}

The Sharpe predictor is a broad concept, and it will give researchers and practitioners the possibility to compare and assess a number of different fairness measures proposed for ML. A good candidate for prediction risk is any {\it subgroup deviation} (see \cite[Section~3.3]{RCWilliamson_AKMenon_2019}). In fact, the Sharpe predictor is meaningful form of doing {\it subgroup aggregation} (see \cite[Section~3.4]{RCWilliamson_AKMenon_2019}).

A key advantage of the Sharpe ratio in portfolio selection is that its computation is robust for an observed series of returns without additional information about their distribution. This aspect can be particular interesting in learning from large datasets with high-variance gradients. The fact that the Sharpe predictor is dimensionless has the potential to introduce novel benchmarking practices when assessing fairness in ML.

The Sharpe ratio is a well-known risk-adjusted performance measure, which enables to evaluate different pairs of return and risk of a portfolio. Another ratio, called information ratio (originally referred to as the appraisal ratio), was introduced in~\cite{JLTreynor_FBlack_1973}. Instead of taking the risk-free return, the information ratio takes a general benchmark return, in the same vein as we did in the experiment of Section~\ref{results_DI}. A negative information ratio is an indication of underperformance relative to the selected benchmark. Also, one can choose any measure of risk replacing the standard deviation of the portfolio return in the denominator of these ratios. There are Sharpe type measures that use Value at Risk (VaR) (called \textit{Reward to VaR}) or Conditional Value at Risk (CVaR) (called \textit{Conditional Sharpe Ratio})~\cite{CBacon_SChairman_2009, VChow_CWLai_2015}. 

\small

\bibliographystyle{plain}
\bibliography{Sharpe-predictor-FML}

\end{document}